# Super-Samples from Kernel Herding

**Yutian Chen**
Department of Computer Science
University of California, Irvine
Irvine, CA 92697

**Max Welling**
Department of Computer Science
University of California, Irvine
Irvine, CA 92697

**Alex Smola**
Yahoo! Research
Santa Clara, CA

## Abstract

We extend the herding algorithm to continuous spaces by using the kernel trick. The resulting "kernel herding" algorithm is an infinite memory deterministic process that learns to approximate a PDF with a collection of samples. We show that kernel herding decreases the error of expectations of functions in the Hilbert space at a rate $\mathcal{O}(1/T)$ which is much faster than the usual $\mathcal{O}(1/\sqrt{T})$ for iid random samples. We illustrate kernel herding by approximating Bayesian predictive distributions.

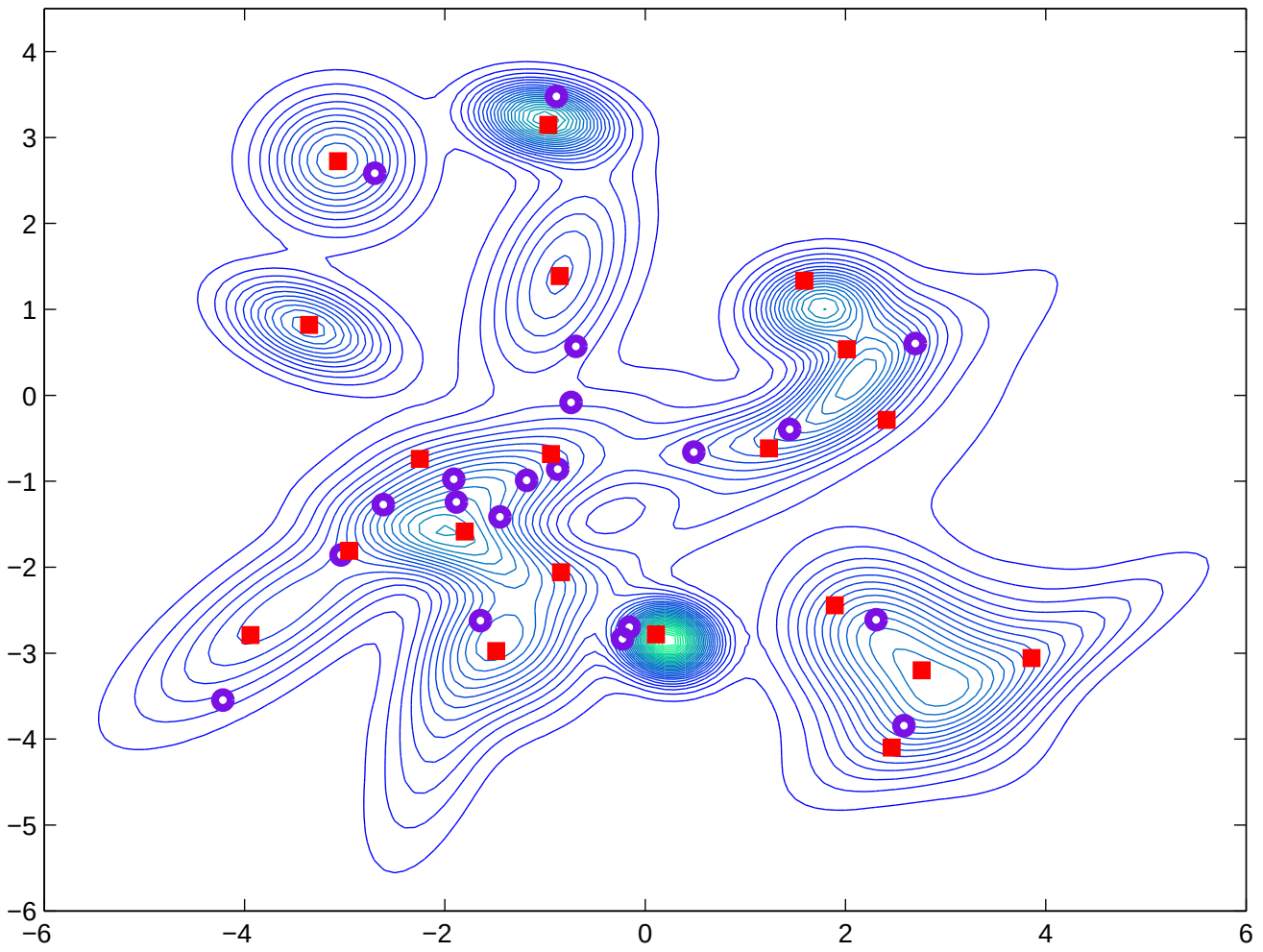

Figure 1: First 20 samples form herding (red squares) versus *i.i.d.* random sampling (purple circles).

## 1 INTRODUCTION

Herding has been understood as a weakly chaotic, nonlinear dynamical system in parameter space, i.e. one can think of it as a mapping $\mathbf{w}_{t+1} = F(\mathbf{w}_t)$ [Welling, 2009a,b, Welling and Chen, 2010, Chen and Welling, 2010]. The discrete states $\mathbf{x}$ play the role of auxiliary variables in this view. However, under this interpretation it has proven difficult to extend herding to continuous spaces. The basic reason is that a finite number of features can not sufficiently control the infinite number of degrees of freedom in continuous spaces leading to strange artifacts in the pseudo-samples[1]. To overcome this we wish to perform herding on an infinite number of features implying the need to switch to a kernel representation.

To achieve that, we will first reinterpret herding as an infinite memory process in the state space $\mathbf{x}$ where we now "marginalize out" the parameters $\mathbf{w}$. Thus, we can consider herding as a mapping $\mathbf{x}_{t+1} = G(\mathbf{x}_1, ..., \mathbf{x}_t, \mathbf{w}_0)$. With two additional very natural assumptions, herding is seen to minimize the squared error between expected feature values evaluated at the true distribution and the empirical distribution obtained from herding. In this new formulation the kernel trick is then straightforward. The main result of this paper is that the error in approximating any function in the RK-Hilbert space defined by the kernel through a Monte Carlo sum decreases as $\mathcal{O}(1/T)$. This is significantly faster than the standard $\mathcal{O}(1/\sqrt{T})$ convergence obtained for iid random samples from $p$. In fact, under the assumption that we perform an unweighed Monte Carlo sum, $\mathcal{O}(1/T)$ convergence is known to be optimal [Kuo and Sloan, 2005]. The reason for the fast convergence is due to negative autocorrelations: the process remembers all previous samples and steers away from regions which have already been (over) sampled. This is illustrated in Figure 1 for a mixture of Gaussians. Similar ideas are the basis for methods such as Quasi Monte Carlo sampling, Quadrature integration and more recently Bayesian integration [Rasmussen and Ghahramani, 2002].

For kernel herding one needs to be able to convolve the density $p$ with the kernel of choice. While this is possi-ble for some rare cases it is hard in general. However, kernel herding can still be very useful if we want to reduce a large collection of samples obtained from a MCMC procedure. Oftentimes, the positive auto-correlations inherent in most MCMC chains are reduced by subsampling. Even in the case when all auto-correlations have been removed this is actually suboptimal, because *negative* auto-correlations may further improve the Monte Carlo approximation. Herding can be used to sub-select a small collection of "super-samples" from a much larger set of MCMC samples. Due to the faster error reduction of herding, in theory we would only need $\sqrt{T}$ samples to obtain the same order of error as $T$ iid random samples. While in practice this is a little optimistic, in our experiments we will show significant boosts in sampling efficiency by using herding.

[1]For instance, herding in a continuous space with features given by the mean and variance will produce two delta-peaks instead of a Gaussian.

We argue that a small collection of super-samples can be beneficial in situations where we wish to average predictions over many predictors. While we may have sufficient time to train up many predictors off-line (through e.g. Bayesian posterior sampling or bagging on bootstrap samples), we may want to be flexible in deciding over how many predictors we average at test time[2]. Herding will precisely organize the samples in an order that is optimal in terms of reducing the error most at every iteration[3]. These ideas are validated with some numerical experiments.

## 2 KERNEL HERDING

We directly describe the herding algorithm in terms of Reproducing Kernel Hilbert Spaces and (potentially) continuous index spaces (note that previous work by [Welling, 2009b,a] cast it in terms of finite-dimensional spaces and discrete domains).

### 2.1 Herding

Let $\mathbf{x} \in \mathcal{X}$ denote some state over an index set $\mathcal{X}$ (typically the space of covariates) and let $\phi : \mathcal{X} \rightarrow \mathcal{H}$ denote a feature map into a Hilbert Space $\mathcal{H}$ with inner product $\langle \cdot, \cdot \rangle$. Given a probability distribution $p(\mathbf{x})$, herding consists of the following update equations for a weight-vector $\mathbf{w} \in \mathcal{H}$

$$\mathbf{x}_{t+1} = \underset{\mathbf{x} \in \mathcal{X}}{\operatorname{argmax}} \langle \mathbf{w}_t, \boldsymbol{\phi}(\mathbf{x}) \rangle \tag{1}$$

$$\mathbf{w}_{t+1} = \mathbf{w}_t + \mathbb{E}_{\mathbf{x} \sim p}[\boldsymbol{\phi}(\mathbf{x})] - \boldsymbol{\phi}(\mathbf{x}_{t+1}) \tag{2}$$

with suitable initialization $\mathbf{w}_0$. We may view this as a weakly chaotic, nonlinear dynamical system over $\mathbf{w}$ [Welling and Chen, 2010, Chen and Welling, 2010] by "maximizing out" the states $\mathbf{x}$. In this case, we may understand herding as taking gradient steps of size 1 on the following (concave, non-positive, scale-free, piecewise linear) function,

$$G(\mathbf{w}) = \Big\langle \mathbf{w}, \underbrace{\mathbb{E}_{\mathbf{x} \sim p}[\boldsymbol{\phi}(\mathbf{x})]}_{:=\mu_p} \Big\rangle - \max_{\mathbf{x} \in \mathcal{X}} \langle \mathbf{w}, \boldsymbol{\phi}(\mathbf{x}) \rangle \tag{3}$$

Here $\mu_p$ denotes the mean operator associated with the distribution $p$ in $\mathcal{H}$, i.e. for $f(\mathbf{x}) = \langle \mathbf{w}, \phi(\mathbf{x}) \rangle$ we have $\mathbb{E}_{\mathbf{x} \sim p}[f(\mathbf{x})] = \langle \mathbf{w}, \mu_p \rangle$. However, we may also take the "dual view" where we remove $\mathbf{w}$ in favor of the states $\mathbf{x}$. This is possible because we can express:

$$\mathbf{w}_T = \mathbf{w}_0 + T\mu_p - \sum_{t=1}^{T} \boldsymbol{\phi}(\mathbf{x}_t) \tag{4}$$

using (2). For ease of intuitive understanding of herding, we temporarily make the assumptions (which are not necessary for proposition 1 to hold):

1. $\mathbf{w}_0 = \mu_p$
2. $\|\boldsymbol{\phi}(\mathbf{x})\|_{\mathcal{H}}^2 = R^2$ for all $\mathbf{x} \in \mathcal{X}$

This condition is easily achieved, e.g. by renormalizing $\phi(\mathbf{x}) \leftarrow \frac{\phi(\mathbf{x})}{\|\phi(\mathbf{x})\|}$ or by choosing a suitable feature map $\phi$ in the first place.

Given the above assumptions and the further restrictions of finite-dimensional discrete state spaces [Welling, 2009b,a], one can show that herding greedily minimizes the squared error $\mathcal{E}_T^2$ defined as

$$\mathcal{E}_T^2 := \left\| \mu_p - \frac{1}{T} \sum_{t=1}^{T} \boldsymbol{\phi}(\mathbf{x}_t) \right\|^2. \tag{5}$$

We therefore see that herding will generate pseudo-samples that greedily minimize this error at every iteration (conditioned on past samples). Note that this does not imply that the total collection of samples at iteration $T$ is jointly optimal. We also note that herding is an "infinite memory process" on $\mathbf{x}_t$ (as opposed to a Markov process) because new samples depend on the entire history of samples generated thus far [Welling and Chen, 2010].

If we manage to find the optimal state $\mathbf{x}_t$ exactly at every iteration then the error in (5) decreases at a rate $\mathcal{O}(T^{-1})$. The proof of this statement follows directly from [Welling, 2009a, Proposition 1 and 2] which was independent of the extra assumptions above. This fast convergence is actually quite remarkable. Note for instance that by generating independent identically distributed *random* samples (iid) from $p$ we get $\mathcal{O}(T^{-\frac{1}{2}})$ convergence while an MCMC method with positive auto-correlation converges even slower. The fact that herding exhibits *faster* convergence can be understood by the fact herding pushes samples away from already explored regions of state space and as such has *negative* auto-correlations. This behavior is reminiscent of Quasi Monte Carlo integration and Bayesian quadrature methods [Rasmussen and Ghahramani, 2002], and is also related to the idea of fast weights for persistent contrastive divergence [Tieleman and Hinton, 2009].

[2]One can imagine a bank trying to present users with personalized ads once they have logged into their website. Depending on the server load the number of predictors used may vary.

[3]Note that this does not imply that the herding set is optimal if we are given the number of samples we are going to be using ahead of time.

### 2.2 Convergence in Hilbert Space

The work of [Welling, 2009b,a] implicitly assumed that there are many more discrete states than features. This has the effect that we only "control" the error in a small subspace of the full state space. The natural question is whether we can *take the (nonparametric) limit where the number of features is infinite*. This is in fact rather straightforward because (1) only depends on the inner product

$$k(\mathbf{x}, \mathbf{x}') := \langle \phi(\mathbf{x}), \phi(\mathbf{x}') \rangle \tag{6}$$

if we plug (4) into (1). This then results in,

$$\mathbf{x}_{T+1} = \tag{7}$$
$$\operatorname*{argmax}_{\mathbf{x} \in \mathcal{X}} \langle \mathbf{w}_0, \phi(\mathbf{x}) \rangle + T\mathbb{E}_{\mathbf{x}' \sim p}[k(\mathbf{x}, \mathbf{x}')] - \sum_{t=1}^{T} k(\mathbf{x}, \mathbf{x}_t)$$

If we initialize $\mathbf{w}_0 = \mu_p$ (Assumption 1), and restrict $\|\phi(\mathbf{x})\| = R$ for all $\mathbf{x} \in \mathcal{X}$ (Assumption 2), the kernel herding procedure becomes:

$$\mathbf{x}_{T+1} = \operatorname*{argmax}_{\mathbf{x} \in \mathcal{X}} \mathbb{E}_{\mathbf{x}' \sim p}[k(\mathbf{x}, \mathbf{x}')] - \frac{1}{T+1} \sum_{t=1}^{T} k(\mathbf{x}, \mathbf{x}_t) \tag{8}$$

and we can see that herding is performing greedy minimization of the error $\mathcal{E}_T$:

$$\begin{aligned} \mathcal{E}_T^2 =& \left\| \mu_p - \frac{1}{T} \sum_{t=1}^{T} \phi(\mathbf{x}_t) \right\|_{\mathcal{H}}^2 \qquad (9) \\ =& \mathbb{E}_{\mathbf{x}, \mathbf{x}' \sim p}[k(\mathbf{x}, \mathbf{x}')] - \frac{2}{T} \sum_{t=1}^{T} \mathbb{E}_{\mathbf{x} \sim p}[k(\mathbf{x}, \mathbf{x}_t)] \\ &+ \frac{1}{T^2} \sum_{t, t'=1}^{T} k(\mathbf{x}_t, \mathbf{x}_{t'}). \end{aligned}$$

The error measures the distance between $p$ and the empirical measure $\hat{p}_T(\mathbf{x}) = \frac{1}{T} \sum_{t=1}^{T} \delta(\mathbf{x}, \mathbf{x}_t)$ given by the herding samples.

This algorithm iteratively constructs an empirical distribution $\hat{p}_T(\mathbf{x})$ that is close to the true distribution $p(\mathbf{x})$. At each iteration, it searches for a new sample to add to the pool. It is attracted to the regions where $p$ is high but repelled from regions where samples have already been "dropped down". The kernel determines how we should measure distances between distributions. Note that for many distributions explicit expressions for $\mathbb{E}_{\mathbf{x}' \sim p}[k(\mathbf{x}, \mathbf{x}')]$ have been obtained. See [Jebara and Kondor, 2003] for details.

The central result of this paper is now that the pseudo-samples generated by kernel herding inherit the fast $\mathcal{O}(T^{-1})$ decrease in error. For a good characterization we need to define the marginal polytope $\mathcal{M}$. It is given by

$$\mathcal{M} := \operatorname{conv} \{ \phi(\mathbf{x}) | \mathbf{x} \in \mathcal{X} \}.$$

It follows that $\mu_p \in \mathcal{M}$ since $\mathcal{X}$ contains the support of $p$. If $\|\phi(\mathbf{x})\| \leq R$ for all $\mathbf{x} \in \mathcal{X}$ it follows that $\|\mu_p\| \leq R$ and consequently by the triangle inequality we have that $\|\mu_p - \phi(\mathbf{x})\| \leq 2R, \forall \mathbf{x}$.

**Proposition 1** *Assume that $p$ is a distribution with support contained in $\mathcal{X}$ and assume that $\|\phi(\mathbf{x})\| \leq R$ for all $\mathbf{x} \in \mathcal{X}$. Moreover assume $\mu_p$ is in the relative interior of the marginal polytope $\mathcal{M}$. Then the error $\mathcal{E}_T$ of (9) will decrease as $\mathcal{O}(T^{-1})$.*

**Proof** We first show that $\|w_t\|$ is bounded for all $t$. For this we introduce the centered marginal polytope

$$\mathcal{C} := \mathcal{M} - \mu_p = \operatorname{conv} \{ \phi(\mathbf{x}) - \mu_p | \mathbf{x} \in \mathcal{X} \}. \tag{10}$$

Using $\mathcal{C}$ the update equations become

$$\mathbf{w}_{t+1} = \mathbf{w}_t - \mathbf{c}_t \text{ where } \mathbf{c}_t := \operatorname*{argmax}_{\mathbf{c} \in \mathcal{C}} \langle \mathbf{w}_t, \mathbf{c} \rangle. \tag{11}$$

This allows us to write the increment in the norm of the parameter vector $\|\mathbf{w}_{t+1}\|$ via

$$\begin{aligned} \|\mathbf{w}_t\|^2 - \|\mathbf{w}_{t+1}\|^2 &= 2 \langle \mathbf{w}_t, \mathbf{c}_t \rangle - \|\mathbf{c}_t\|^2 \qquad (12) \\ &\geq 2 \|\mathbf{c}_t\| \left[ \|\mathbf{w}_t\| \left\langle \frac{\mathbf{w}_t}{\|\mathbf{w}_t\|}, \frac{\mathbf{c}_t}{\|\mathbf{c}_t\|} \right\rangle - R \right] \end{aligned}$$

The inequality follows from $\|\mathbf{c}_t\| \leq 2R$. If we can show

$$\left\langle \frac{\mathbf{w}_t}{\|\mathbf{w}_t\|}, \frac{\mathbf{c}_t}{\|\mathbf{c}_t\|} \right\rangle =: \gamma_t \geq \gamma^* > 0 \tag{13}$$

for all $\mathbf{w}$ then it follows immediately that $\|\mathbf{w}\| \leq R/\gamma^*$: in this case we have $\|\mathbf{w}\| \gamma_t - R \geq (R/\gamma^*)\gamma^* - R = 0$.

To see (13) recall that $\mu_p$ is contained inside the relative interior of $\mathcal{M}$, i.e. there exists an $\epsilon$-ball around $\mu_p$ that is contained in $\mathcal{M}$. Consequently $\gamma^* \geq \epsilon$.

Since $\|\mathbf{w}_t\| = \left\| w_0 + T\mu_p - \sum_{t=1}^{T} \phi(\mathbf{x}_t) \right\| \leq R/\gamma^*$ it follows by dividing by $T$ that

$$\left\| \mu_p - T^{-1} \sum_{t=1}^{T} \phi(\mathbf{x}_t) \right\| \leq T^{-1}[\|w_0\| + R/\gamma^*]. \tag{14}$$

This proves the claim of $\mathcal{O}(T^{-1})$ convergence to $\mu_p$. ■

The requirement that $\mu_p \in \mathcal{M}$ is easy to check: it occurs whenever $p$ has full support with respect to the domain of optimization (provided that $\phi(\mathbf{x})$ is characteristic and therefore leads to unique representations).

**Corollary 2** *Herding converges at the fast rate even when (1) is only carried out with some error provided that we obtain samples* $\mathbf{x}_{t+1} \in \mathcal{X}$ *which satisfy*

$$\left\langle \frac{\mathbf{w}_t}{\|\mathbf{w}_t\|}, \frac{\phi(\mathbf{x}_{t+1}) - \mu_p}{\|\phi(\mathbf{x}_{t+1}) - \mu_p\|} \right\rangle \geq \bar{\rho} > 0 \quad (15)$$

This condition is reminiscent of Boosting algorithms where the weak learner is not required to generate the optimal solution within a given set of hypotheses but only one that is sufficiently good with regard to a nonzero margin. It is also related to the *perceptron cycling theorem* [Block and Levin, 1970] where $\mathcal{X}$ is assumed to have finite cardinality but which guarantees convergence even when $\bar{\rho} = 0$.

We can allow $\mu_p$ to lie on a facet of $\mathcal{M}$ in which case we have the following corollary.

**Corollary 3** *Whenever* $\mu_p$ *lies on a facet of the marginal polytope* $\mathcal{M}$ *it suffices that we restrict ourselves to optimization over the vertices generating the facet. In this case,* $\mu_p$ *lies within the relative interior of the now restricted polytope.*

We finally want to show that the $\mathcal{O}(T^{-1})$ convergence of the error $\mathcal{E}_T$ as proved above implies that the error of any integral over a function in our RKHS will also converge at the same fast rate:

**Proposition 4** *For any* $f \in \mathcal{H}$*, the error* $|\mathbb{E}[f]_p - \mathbb{E}[f]_{\hat{p}_T}|$ *will decrease as* $\mathcal{O}(T^{-1})$*. Moreover this condition holds uniformly, that is* $\sup_{\|f\|\leq 1} |\mathbb{E}[f]_p - \mathbb{E}[f]_{\hat{p}_T}|$ *also decreases at rate* $\mathcal{O}(T^{-1})$*.*

To prove this we will need the following lemma,

**Lemma 5 (Koksma Hlawka Inequality)** *For any* $f \in \mathcal{H}$ *we have*

$$|\mathbb{E}[f]_p - \mathbb{E}[f]_{\hat{p}_T}| \leq \|f\|_{\mathcal{H}} \|\mu_p - \mu_{\hat{p}_T}\|_{\mathcal{H}} \quad (16)$$

The above inequality is the simply a consequence of the Cauchy Schwartz inequality. It is known as the Koksma-Hlawka inequality in the analysis of Quasi Monte Carlo methods. Clearly, with this lemma proposition 4 follows. In fact, this technique was used by [Song et al., 2008] in the context of density estimation. The key novelty in the present paper is that we have a *simple* and *explicit* algorithm for obtaining fast rates of approximation which are considerably better than the $\mathcal{O}(T^{-\frac{1}{2}})$ rates usually available via sampling.

For some special kernel functions, we can get better properties for the samples generated by herding. Since the error in (9) converges to 0, following Lemma 4 in [Gretton et al., 2008], we know the maximum mean discrepancy (MMD) on the unit ball of $\mathcal{H}$ also converges to 0. If the RKHS $\mathcal{H}$ is universal, combining with Theorem 3 in the same paper, it suggests that the probability distribution of herding samples $\hat{p}_T$ converges to the true distribution $p$ at rate $\mathcal{O}(T^{-1})$ as $T \to \infty$. Examples of kernels with universal RKHS are Gaussian and Laplace kernels defined on a compact space.

**Corollary 6** *An active learning algorithm selecting labels in accordance with the herding algorithm has guaranteed rate of convergence in terms of its bias of* $\mathcal{O}(T^{-1})$*. Moreover, the submodular greedy algorithm of [Guestrin et al., 2005] has therefore also at least the same approximation rate since it is within a constant fraction* $(1 - e^{-1})$ *of optimality.*

In summary, kernel herding generates samples that are much more informative than iid samples: for every $n$ herding samples we will need $O(n^2)$ iid samples to achieve the same error reduction. For this reason we will call herding samples *super-samples* from now on.

## 3 Experiments

In this section, we want to show that herding is able to draw better samples than random sampling from the true distribution. We first illustrate the behavior of herding on low dimensional synthetic data, compare the approximation of integrals between the super samples and iid samples, and then we show an application where we compress the size of a collection of posterior samples required for computing the predictive probability of Bayesian logistic regression.

### 3.1 Synthetic Data

#### 3.1.1 Matching the True Distribution

We first visualize herding on a 2-D state space. We randomly construct a 2 dimensional Gaussian mixture (GM) model with 20 components whose equiprobability contours are shown in Figure 1. With a Gaussian kernel, the integral in (8) can be analytically calculated implying that we can run herding directly on the GM distribution.

A few random samples are first drawn to provide reasonable seeds for the maximization. Then, we sequentially generate super-samples by (8). At each iteration, starting from the best auxiliary sample that maximizes (8), we run a gradient ascent algorithm to obtain a new super sample. Figure 2 shows the linear increase of $1/\mathcal{E}_T$ as a function of $T$.

In Figure 1, the first 20 super samples are plotted in comparison with 20 iid samples from the GM model. For iid samples, due to the inherent randomness, some modes receive too many points while others get too few or even none. In contrast, the samples from herding always try to repel from each other and are distributed optimally (given earlier samples) to represent the density function.

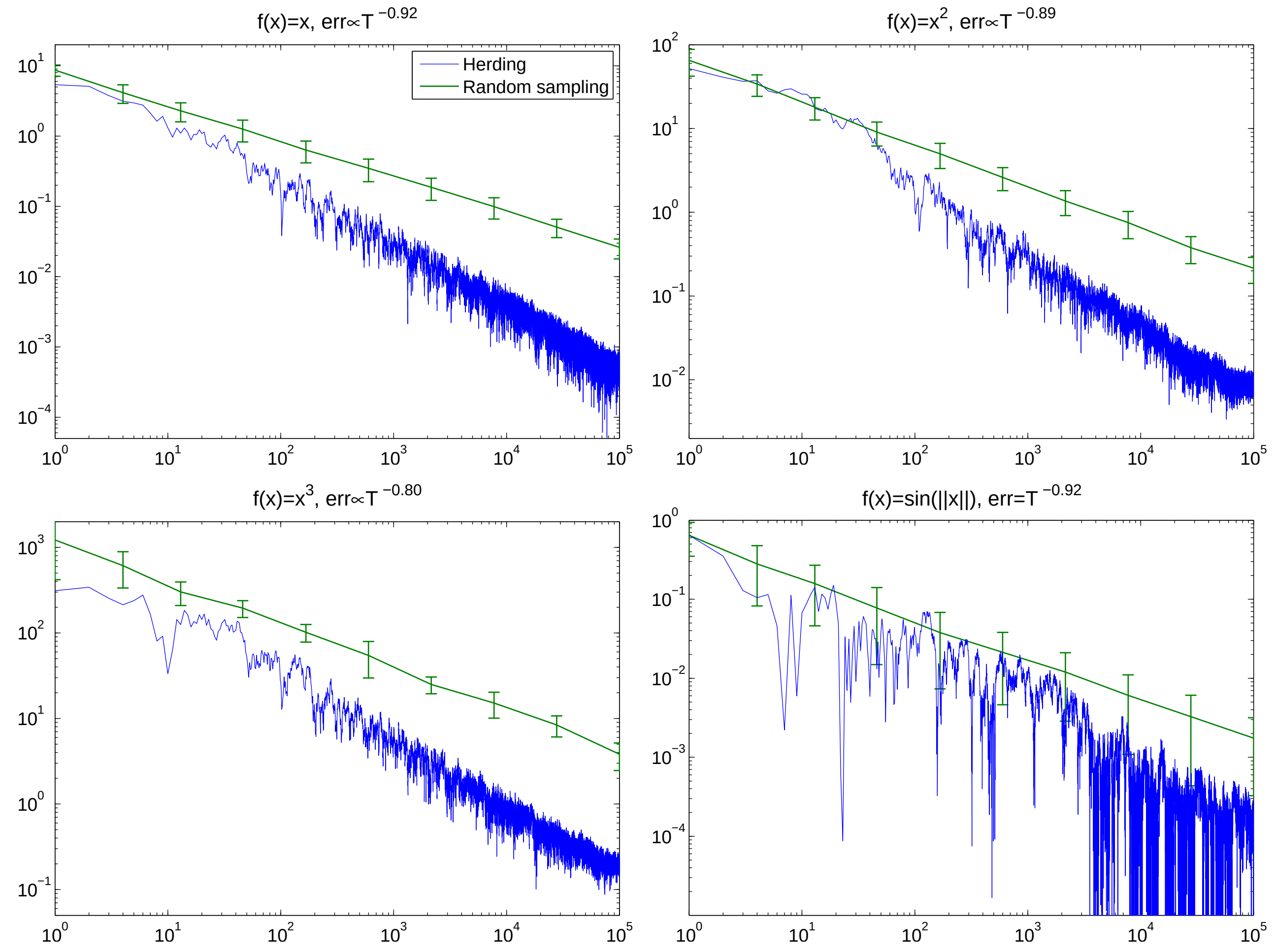

Figure 3: Error in estimating the expectation of four functions, by herding (blue) and random sampling (green) as a function of the number of samples. The decreasing speed of the upper bound of the error is shown on top of each figure.

Since the expectation of any function in a model can be approximated by summation over its samples, we are interested in how well the super samples can be used to estimate these averages. We generate a 5 dimensional GM model with 100 components as the target distribution $p$. We compute the error of the expectation on four functions: the first three moments, and a nonlinear function. For the $m$'th moment, $m = 1, 2, 3$, we first calculate the average of $x_{i,t}^m$ over $t$ (the index of herding samples) in each dimension as a function of $T$ (the number of super samples). Then the RMSE of the estimated moments over all the dimensions is computed as

$$\text{err}(\mathcal{S}_T) = \left( \frac{1}{d} \sum_{i=1}^{d} (\langle x_i^m \rangle_{\mathcal{S}_T} - \langle x_i^m \rangle_p)^2 \right)^{\frac{1}{2}} \tag{17}$$

For the fourth function, we use a sine of the norm of a point: $f(x) = \sin \|x\|$. In comparison, we compute the mean and standard deviation of the errors obtained by a set of random samples as the benchmark. The results are shown in Figure 3 with their estimated convergence rates. The error of approximation by herding is much smaller than random sampling with the same number of points for all the 4 functions, also their convergence rates are close to the theoretical value $\mathcal{O}(T^{-1})$.

### 3.1.2 Matching empirical distribution

When the integration in (8) can't be computed analytically, it would be difficult to run herding to accurately match the true distribution especially in high dimensional spaces. However, if we have a set of random samples, $\mathcal{D}$, from the distribution, it is straightforward to run herding to match the empirical distribution. We can thereby represent the true distribution by the super samples $\mathcal{S}$ with the same accuracy as $\mathcal{D}$ but with many fewer samples. A set of $10^5$ iid samples is drawn from a 5-D GM model, and then herding is run taking $\mathcal{D}$ as the true distribution. Since in this case $p$ in the (8) is taken to be the empirical distribution, the integral is simply a summation over all the points in $\mathcal{D}$.

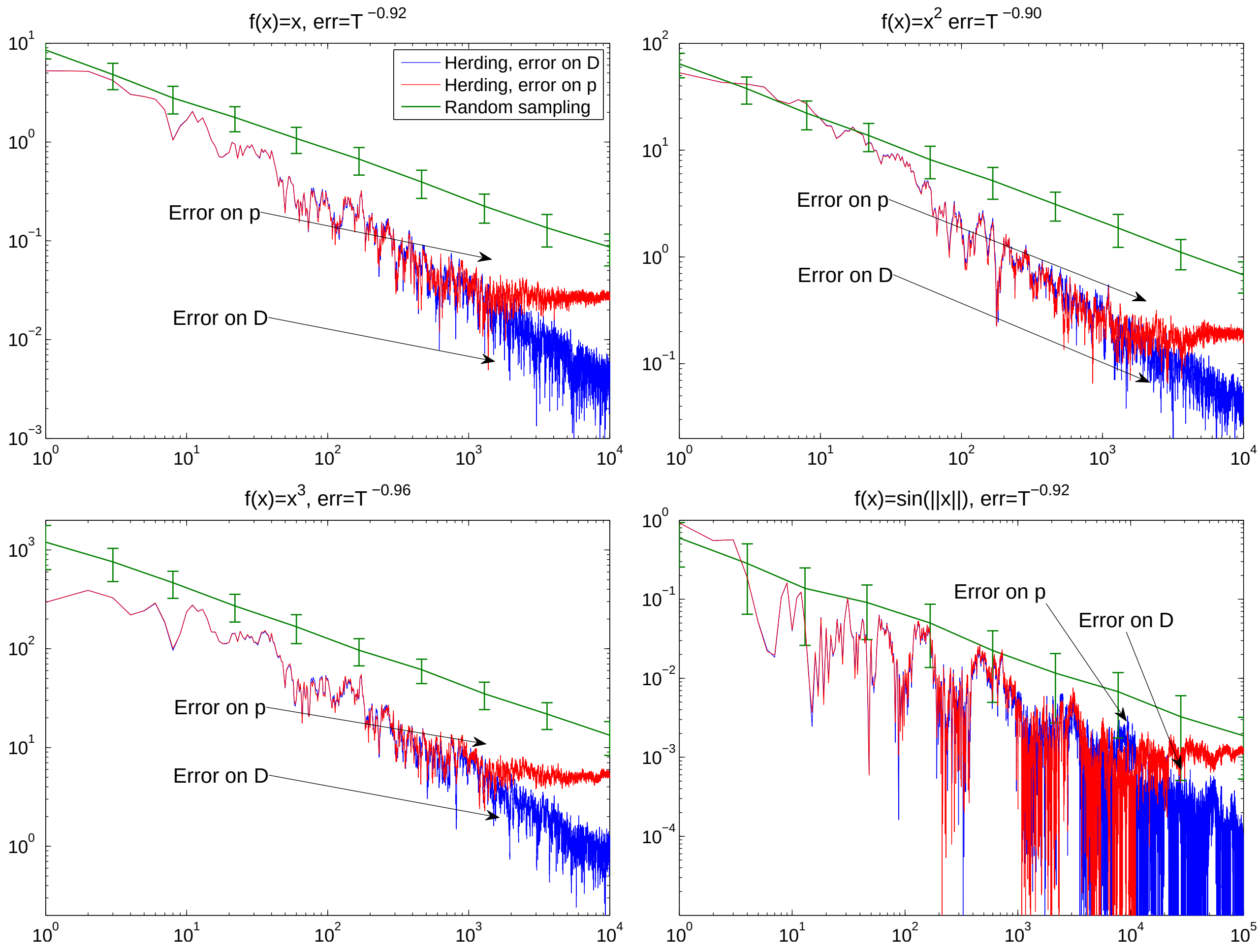

Figure 4: Error in estimating the expectation of four functions by herding on the true distribution $p$ (red) and the empirical distribution $\mathcal{D}$ (blue) as a function of the number of samples. The convergence rate of the error on $\mathcal{D}$ (measured as slope of the upper bound of the herding error) is shown on top of each figure. The error of random sampling on $p$ (green) is also plotted for comparison.

We again compare the estimation of function expectations between herding and random samples. However, this time we can compute two errors, one on the empirical distribution $\mathcal{D}$ and the other on the true distribution $p$. Since the distribution of $\mathcal{S}$ will converge to the empirical distribution, the error between $\mathcal{S}$ and $\mathcal{D}$ will keep decreasing as in Figure 3 while the error between $\mathcal{S}$ and $p$ will not. Instead, it will converge to the error incurred by the empirical distribution relative to $p$ and this is the point where the set $\mathcal{S}$ is large enough to replace $\mathcal{D}$. We can find from Figure 4 that for $10^5$ iid samples, we only need at most $2000$ super samples for the first three functions, and $10^4$ for the last function to achieve similar precision. This is a significant reduction whenever evaluating $f$ is expensive, e.g. for user interaction data.

### 3.2 Approximating the Bayesian Posterior

Next we consider the task of approximating the predictive distribution of a Bayesian model. Alternatively, this idea can be applied to find a small collection of good predictive models to be used in bagging. Assume we have drawn a large number of parameters, $\mathcal{D}$, using MCMC from the posterior distribution (or we have learned a large number of predictors on bootstrap samples). For reasons of computational efficiency, we may not want to use all the samples at test time. One choice is to down-sample the MCMC chain by randomly sub-sampling from $\mathcal{D}$. Another choice is to run herding on the empirical distribution. With the convergence property on any function in the Reproducing Kernel Hilbert Space, we know that prediction by $\mathcal{S}$ will converge to that by $\mathcal{D}$. Furthermore, we can get a significant speed up with a few super samples during the prediction phase without much loss of accuracy.

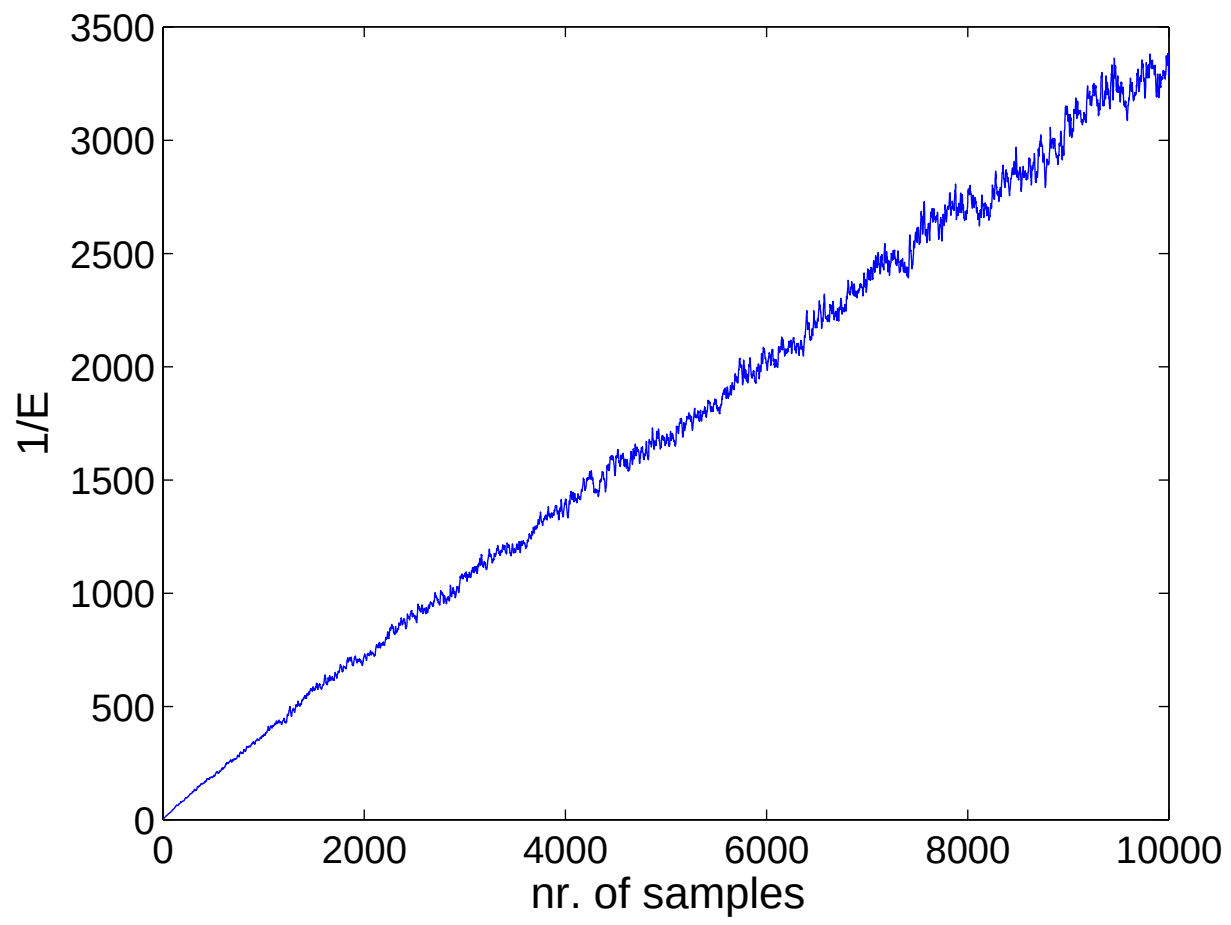

Figure 2: Linear relationship between $1/\mathcal{E}_T$ and $T$

We use the spambase data set from the UCI machine learning repository[4] for the experiment, which has 4601 instances with 57 real attributes and 1 binary class label. The data set is split into a training set of 3000 data points and a test set of 1601 data points. A logistic regression model is built with a Gaussian prior on the weights $\theta$. The training set is whitened by PCA and then fed into the model to draw posterior samples by the Metropolis-Hasting algorithm with a Gaussian proposal distribution. The resulting set $\mathcal{D}$ consists of $10^5$ samples sub-sampled by a factor of $100$ from the Markov chain to reduce the autocorrelation. We whiten $\mathcal{D}$ using PCA and run herding on this empirical distribution with an isotropic Gaussian kernel with $\sigma = 10$. This is equivalent to run herding on the original parameter set with a Gaussian kernel whose covariance matrix is a multiple of the covariance matrix of $\mathcal{D}$. At each iteration, we use the sample of $\theta$ from $\mathcal{D}$ that maximizes (8) as a new super sample, without any further local maximization. This corresponds to running herding in the discrete domain, $\mathcal{X} = \mathcal{D}$, and all the theoretical conclusions in section 2 also apply to this case.

We compare the predictions made by $\mathcal{S}$ with those made by the whole set $\mathcal{D}$ on the test data. Figure 5 shows the RMSE of the predictive probability by herding over all the test data points as a function of the number of super samples.

$$\mathrm{RMSE}^2(\mathcal{S}_T, \mathcal{D}) \tag{18}$$

$$= \frac{1}{N}\sum_{n=1}^{N}\left[\frac{1}{T}\sum_{t=1}^{T} p(y_n|x_n,\theta_t) - \frac{1}{|\mathcal{D}|}\sum_{i=1}^{|\mathcal{D}|} p(y_n|x_n,\theta_i)\right]^2$$

For comparison, we randomly draw a subset of $\mathcal{D}$ by bootstrap sampling and compute the error in the same way (the performance of down-sampling the Markov chain or randomly sampling without replacement is very similar to random sampling, and is thus not shown in the figure). We can easily observe the advantage of herding over random sampling. The error of herding decreases roughly as $\mathcal{O}(T^{-0.75})$, while the error of random sampling decreases as $\mathcal{O}(T^{-0.5})$.

[4] http://archive.ics.uci.edu/ml/

Now we'd like to estimate how many super samples are needed to achieve the same precision as $\mathcal{D}$ on the true posterior. Assume for now that the samples in $\mathcal{D}$ are iid. Then the average predictive probability $p(x_n|y_n,\mathcal{D}) = \frac{1}{|\mathcal{D}|}\sum_{i=1}^{|\mathcal{D}|} p(y_n|x_n,\theta_i)$ is the average of $|\mathcal{D}|$ independent, unbiased estimates. Since we can compute the standard deviation of these estimates on $\mathcal{D}$, the standard deviation of the average predictive probability becomes $\mathrm{std}(p(y_n|x_n,\mathcal{D})) = \mathrm{std}(p(y_n|x_n,\theta_i))/\sqrt{|\mathcal{D}|}$, and then its mean over all test data points gives an estimate to the standard deviation of the error in general, which is the dashed line in Figure 5.

We can decompose the error of herding on the true posterior

$$\begin{aligned}\mathrm{RMSE}(\mathcal{S}_T, p) &\le \mathrm{RMSE}(\mathcal{S}_T, \mathcal{D}) + \mathrm{RMSE}(\mathcal{D}, p) \\ &\approx \mathrm{RMSE}(\mathcal{S}_T, \mathcal{D}) + \overline{\mathrm{std}(p(y_n|x_n,\mathcal{D})}.\end{aligned}$$

When the first term is smaller than the second term, the error of herding mainly comes from the error of $\mathcal{D}$, and we can claim that more herding samples will not improve the prediction much. Since the MCMC samples in $\mathcal{D}$ are not independent, the error of $\mathcal{D}$ can only be larger than the estimated value, and we'll need even fewer samples to reach the same accuracy. In our experiment, for a set of $10^5$ samples, we only need 7000 super samples.

In fact, we have drawn another much larger set of $2.5 \times 10^6$ posterior samples, $\tilde{p}$, and estimate the error of $\mathcal{S}$ on $p$ by $\mathrm{RMSE}(\mathcal{S}, \tilde{p})$ (the red line in Figure 5). We find that the line starts to level off with even fewer (about 3000) super samples and the converged value equals $\mathrm{RMSE}(\mathcal{D}, \tilde{p})$. In summary, we can compress the set of parameters by $93\%$ or $97\%$.

In Figure 6, we show the classification accuracy of herding on the test set. In comparison, we also draw the accuracy of the whole sample set (red), and 10 random subsets of $\mathcal{D}$. The prediction of herding converges fast to that of $\mathcal{D}$ which is considered ground truth for the herding algorithm. In contrast, the prediction made by random subsets fluctuates strongly. In particular, we only need about 20 super-samples to get the same accuracy as $\mathcal{D}$, while we need about 200 random samples.

# 4 DISCUSSION

Kernel herding extends the original herding algorithm to continuous spaces, and generates samples that contain more information than IID samples. For a few distributions on which we can compute the convolution between $p$ and the kernel, herding samples approximate the expectation of any function in the Hilbert space at a rate of $\mathcal{O}(T^{-1})$, much faster than the Monte Carlo method. For other distributions, given a collection of samples, KH filters out part of its inherent randomness, and converts it into a much more compact set of super samples with the same accuracy as the empirical distribution.

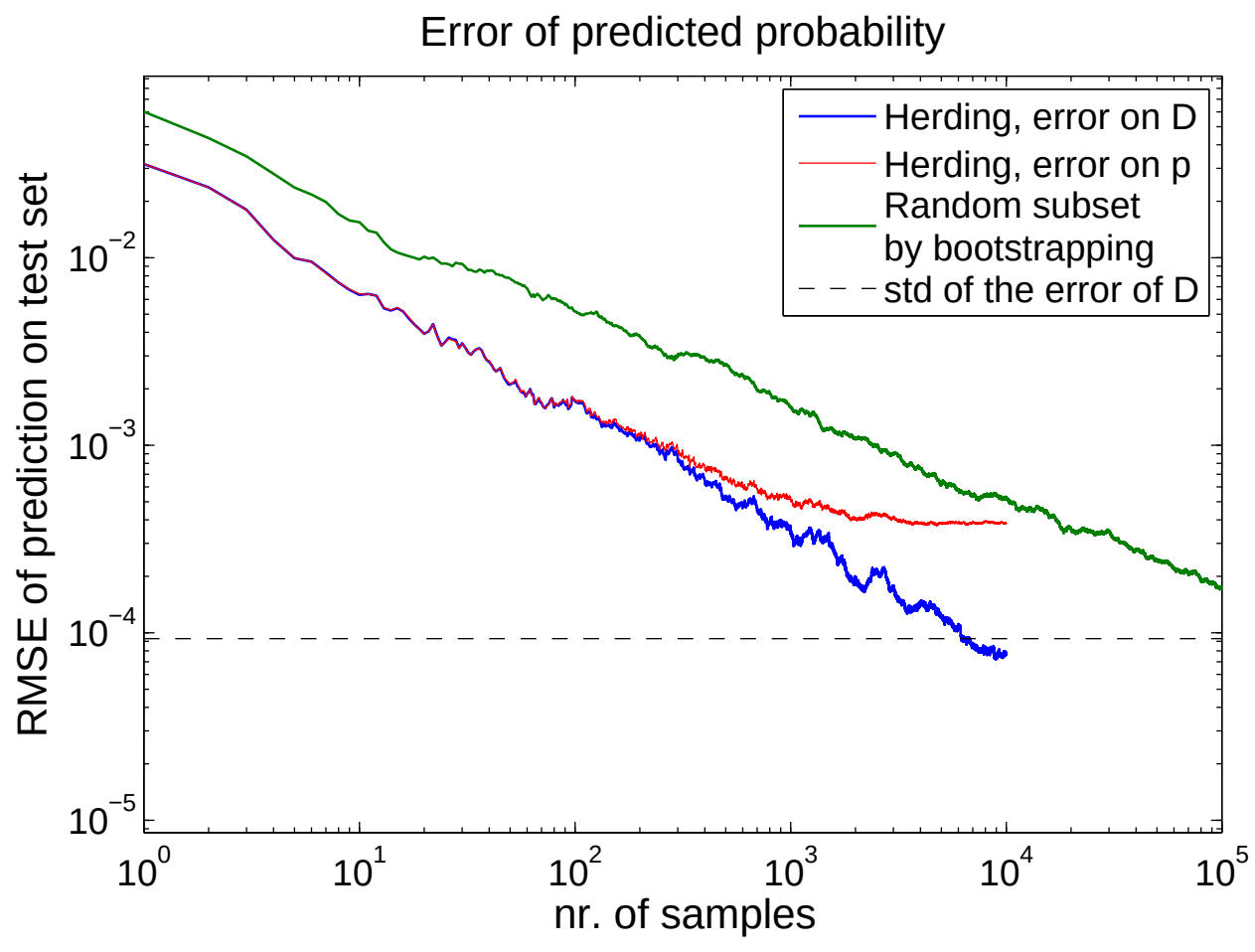

Figure 5: RMSE of the predicted probability of herding (blue) and a random subset (blue) w.r.t. the whole sample set.

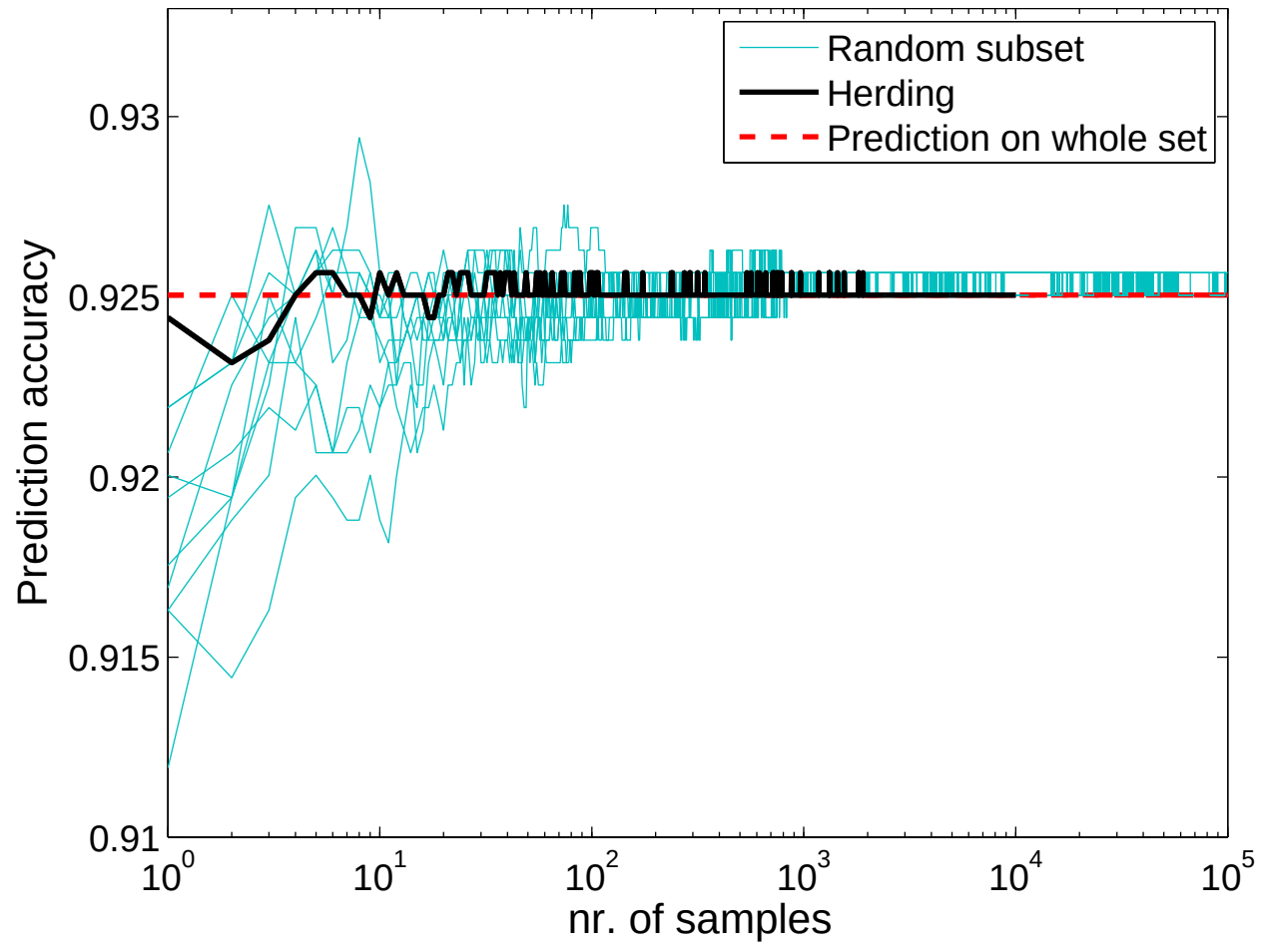

Figure 6: Prediction accuracy of herding (black), 10 random subsets (cyan) and the whole sample set (red).

Despite the power of KH already shown in this paper, we only use a Gaussian kernel without utilizing much information about the function of interest. If we already know the function we want to integrate over or a distribution of functions, it will be possible to design a better kernel that minimizes the expected error w.r.t. to that distribution. This is a promising future research direction.

Also, the idea of repelling samples from those areas that have been explored is not only useful for herding. Incorporating the negative auto-correlation between samples to MCMC or other methods should help speed up mixing. And in the other direction, introducing stochastic methods to approximate the convolution in (8) should make KH more practical in applications where general distributions $p$ are required.